\documentclass[11pt,letterpaper]{article}
\usepackage{naaclhlt2016}
\usepackage{graphicx}
\usepackage{booktabs}
\usepackage{url}
\usepackage{xcolor}
\usepackage{floatrow}
\usepackage[hyperindex,breaklinks]{hyperref}
\PassOptionsToPackage{linktocpage}{hyperref}

\DeclareFloatFont{tiny}{\footnotesize}% "scriptsize" is defined by floatrow,"tiny" not
\naaclfinalcopy % Uncomment this line for the final submission
\title{UH-PRHLT at SemEval-2016 Task 3:\\Combining Lexical and Semantic-based Features for\\Community Question Answering}

 \author{Marc Franco-Salvador$^{1}$, Sudipta Kar$^{2}$, Thamar Solorio$^2$, \and Paolo Rosso$^{1}$\\
 $^1$ Pattern Recognition and Human Language Technology (PRHLT) research center\\Universitat Polit{\`e}cnica de Val{\`e}ncia, Spain\\
 $^2$ Department of Computer Science\\University of Houston, United States\\
 {\tt mfranco@prhlt.upv.es, skar3@uh.edu,}\\
 {\tt tsolorio@uh.edu, prosso@prhlt.upv.es}
}

\begin{document}
\maketitle
\begin{abstract}

%The SemEval 2016 Task 3 on \textit{Community Question Answering} aims at covering a full task on Community Question Answering by means of four subtasks: i)
%English question-comment similarity ranking; ii) English question-related question similarity ranking; iii) English question-external comment similarity ranking; and iv) Arabic 
%question-related question with correct answer re-ranking. 
In this work we describe the system built for the three English subtasks of the SemEval 2016 Task 3 by the Department of Computer Science of the University of Houston (UH) and the Pattern Recognition and Human Language Technology (PRHLT) research center - Universitat
Polit{\`e}cnica de Val{\`e}ncia: UH-PRHLT.
Our system represents instances by using both lexical and semantic-based similarity measures between text pairs. Our semantic features include the use of distributed 
representations of words, knowledge graphs generated with the BabelNet multilingual semantic network, and the FrameNet lexical database.  
Experimental results outperform the random and Google search engine baselines in the three English subtasks. Our approach obtained the highest results of subtask B compared to the other task participants.

\end{abstract}

\section{Introduction}
\label{intro}

The key role that the Internet plays today for our society benefited the dawn of thousands of new Web social activities. Among those, forums emerged with special relevance 
following the paradigm of the Community Question Answering (CQA). These type of social networks allow people to post a question to other users of that 
community. The 
usage is simple, without much restrictions, and infrequently moderated. The popularity of CQA is a strong indicator that users receive some good and valuable answers. However, there are several issues related to that 
type of community. First is the large amount of answers received that makes it difficult and time-consuming for users to search and distinguish the good ones. This is 
exacerbated with the amount of noise that these questions contain. It is not uncommon to have wrong or misguiding answers that produce more unrelated answers, discussions and 
sub-threads. Finally, there is a lot of redundancy, questions may be repeated or closely related to previously asked questions.

Details of the SemEval 2016 Task 3 on CQA can be found in the overview paper~\cite{nakov-EtAl:2016:SemEval}. 
In this work we evaluate the three English-related Task 3 subtasks on CQA. We first represent each instance to rank --- question versus (vs.) comments, question 
vs. related questions, or question vs. comments of related questions  --- with a set of similarities computed at two different levels: lexical and semantic. This representation 
allows 
us to estimate the relatedness between text pairs in terms of what is explicitly stated and what it means. 
Our lexical similarities employ representations such as word and character $n$-grams, and bag-of-words (BOW). The semantic similarities include the use of distributed word 
bidirectional alignments, distributed representations of text, knowledge graphs, and frames from the FrameNet lexical database~\cite{baker1998berkeley}. 
This type of dual representations have been successfully employed for question answering by the highest performing system in the previous edition of this SemEval 
task~\cite{tran2015jaist}. Other Natural Language Processing (NLP) tasks such as cross-language document retrieval and categorization also benefited from similar 
representations~\cite{francosalvador-rosso-navigli:2014:EACL}. In this task, if the question or comment includes multiple text fields, e.g. body and subject, similarities are 
estimated using all possible combinations (see Section~\ref{subtaskRepr}). Finally, the ranking of instances is performed using a state-of-the-art machine-learned ranking 
algorithm: SVM$_{rank}$.
 
%The rest of the paper is structured as follows. Section~\ref{relWork} overviews the related work on CQA. Next, Section~\ref{approach} details our approach for the SemEval 2016
%Task 3 on CQA. Section~\ref{eval} presents our evaluation and analysis. Finally, in Section~\ref{conclu} we provide our conclusions and discuss future work. Additional analysis 
%and comparison with the other submitted systems are available in the SemEval 2016 Task 3 overview~\cite{nakov-EtAl:2016:SemEval}.

\section{Related Work}
\label{relWork}
%\textcolor{red}{Marc consider discussing only top two systems at SemEval 2015 and just a couple of sentences for all other related work not an entire paragraph.}
Automatic question answering has been a popular interest of research in NLP from the beginning of the Internet to more recently where voice interfaces have been 
incorporated~\cite{rosso2012voice}. The use of BOW representations allowed to 
correctly answer 60\% of the questions of the first large-scale question answering evaluation at the TREC-8 Question Answering 
track~\cite{voorhees1999trec}.  More complex systems used inference rules to connect expressions between questions and answers~\cite{lin2001discovery}. 
Similarly, \newcite{ravichandran2002learning} employed bootstrapping to generate surface text patterns in order to successfully answer questions. 
Other works such as~\newcite{buscaldi2010answering} are based on the redundancy of $n$-grams in order to find one or more text fragments that include tokens of the original 
question and the answer.
\newcite{jeon2005finding} studied the semantic relatedness between texts for question answering. They used translation obfuscation to paraphrase the text 
and to detect which terms are closer in context. Probabilistic topic models have been also useful for detecting the semantics in this task. \newcite{celikyilmaz2010lda} used 
Latent Dirichlet Allocation (LDA)~\cite{blei2003latent} for representing questions by means of latent topics.

The previous edition of the SemEval CQA task included two English subtasks~\cite{nakov2015semeval}. The first one was focused on classifying answers as 
\textit{good}, \textit{bad}, or \textit{potentially relevant} with respect to one question. The second subtask answered a question with \textit{yes}, \textit{no}, or 
\textit{unsure} based on the list of all answers. In addition, the first subtask was also available in Arabic. Several teams experimented with complex 
solutions that included meta-learning, external resources, and linguistic features such as syntactic relations and distributed word representations. Similarly to our work, the highest performing approach employed a combination of lexical and semantic-based similarity measures~\cite{tran2015jaist}. Its semantic features included the use of 
probabilistic topic models, translation obfuscation-based alignments, and pre-computed distributed representations of words both generated with the 
word2vec\footnote{\url{https://code.google.com/archive/p/word2vec/}} and GloVe\footnote{\url{http://nlp.stanford.edu/projects/glove/}} toolkits. Their lexical features included 
BOW, word alignments, and noun matching. They employed a regression model for classification. Another interesting approach, \newcite{hou2015hitszicrc}, included textual features 
--- 
word lengths and punctuation --- in addition to syntactical-based features --- Part-of-Speech (PoS) tags.

In this work we aim at differentiating from the other approaches by enhancing our ranking model with new similarity measures. These include the use of knowledge graphs obtained 
using 
the largest multilingual semantic network --- BabelNet --- frames from the FrameNet lexical database, and bidirectional distributed word alignments.

\section{Lexical and Semantic-based Community Question Answering}
\label{approach}

In this section we detailed the system that we designed for this CQA task. First in
%Section~\ref{preproc} we describe the pre-processing that we followed. Next, in 
Section~\ref{feats} we described our set of lexical features and semantic-based ones. Next, in Section~\ref{subtaskRepr} we detail the specific adaptation that we employed 
for each subtask and the ranking algorithm that we used. We note that all our features are similarity scores obtained with different text similarity measures. More details and
examples can be found in their respective papers.

%\subsection{Data Pre-processing}
%\textcolor{red}{Consider removing this single para section and move this text to dataset and methodology}
%\label{preproc}
% I suggest we remove the bits of text explaining the organization of the paper. A good paper organization should not require this
%We removed stopwords, and employed lemmatisation and stemming. However, for the distributed representation and knowledge graph-based features we did not employ stemming. 
%These decisions were motivated for performance reasons during our prototyping.

\subsection{Feature Description}
\label{feats}

Our system exploits both the verbatim and the contextual similarities between texts, i.e., questions and comments. In Section~\ref{lexical} we detailed our lexical and in 
Section~\ref{semantic} our semantic-based features.

\subsubsection{Lexical Features}
\label{lexical}
%The lexical features that we employed are the following:
%No need for the text below since this is explained in section 3.3
%\textcolor{red}{We have used several lexical features and they were extracted in multiple levels. For example, in the question similarity task, similarity was calculated between subjects of two questions, bodies of two questions and subject plus body of two questions.}
The lexical features that we employed are the following:
\begin{itemize}
  \item \textcolor{black}{
	\textbf{Cosine Similarity.} We used cosine similarity to measure lexical similarity between two text snippets. We calculated cosine similarity based on word 
$n$-grams(n=1,2), character 3-grams and tf-idf~\cite{Salton:1986:IMI:576628} scores of words.}
    
    \item \textcolor{black}{
	\textbf{Word Overlap.} We used the count of common words between two texts. This count was normalized by the length.}

    \item \textcolor{black}{
	\textbf{Noun Overlap.} We used NLTK\footnote{\url{http://www.nltk.org/}} to part-of-speech tag the text and computed the normalized count of overlapping nouns in two texts 
as a similarity measure.}
    
    \item \textcolor{black}{
	\textbf{N-gram Overlap.} We computed the normalized count of common $n$-grams($n$=1,2,3) between two texts.}

\end{itemize}

\subsubsection{Semantic Features}
\label{semantic}

The semantic features that we employed are the following:

\begin{itemize}
  \item \textbf{Distributed representations of texts.} We used the continuous Skip-gram model~\cite{mikolov2013distributed} of the word2vec toolkit to generate distributed 
representations of the words
of the complete English Wikipedia.\footnote{We used 200-dimensional vectors, context windows of size 10, and 20 negative words for each sample.} Next, for each text, e.g. question 
or comment, we averaged its word vectors in order to have a single representation of its content as this setting has shown good results in other NLP tasks
(e.g. for language variety identification~\cite{FrancoSalvadorCLEF2015} and discriminating similar languages~\cite{FrancoSalvadorDSL2015}).
%Work by ~\cite{le2014distributed} also proved that the representation is a competitive alternative to distributed representations of sentences. 
Finally, the similarity between texts, e.g. question vs. comment, is estimated using the cosine similarity.
  \item \textbf{Distributed word alignments.} The use of word alignment strategies has been employed in the past for textual semantic relatedness~\cite{hassan2011semantic}. 
  \newcite{tran2015jaist} employed distributed representations to align the words of the question with the words of the comment. A more recent work introduced the Continuous Word 
Alignment-based Similarity Analysis (CWASA)~\cite{FrancoSalvador2016:KNOSYS}. CWASA uses distributed representations to measure the similarity by double-direction aligning words 
of texts. In this work we selected as feature the similarity provided by CWASA between questions and comments.
   \item \textbf{Knowledge graphs.} A knowledge graph is a labeled, weighted, and directed graph that expands and relates the concepts belonging to a text. Knowledge Graph
Analysis (KGA)~\cite{FrancoSalvadorIPM2016} measures semantic relatedness between texts by means of their knowledge graphs. In this work we used the 
BabelNet~\cite{navigli2012babelnet} multilingual semantic network to generate knowledge graphs from questions and comments, and measured their similarity using KGA.
 \item \textcolor{black}{
	\textbf{Common frames.} We used Framenet ~\cite{baker1998berkeley} to extract the frames associated with the lexical items in the text. For each frame present in the text, we calculated the common lexical items between sentences associated with this frame. The goal is to allow inference of similarity at the level of semantic roles.}
\end{itemize}

As additional feature, for Subtasks A and C we also used the ranking provided by the Google search engine for the questions related to the original questions.

\subsection{Data Representation and Ranking}
\label{subtaskRepr}

Due to the representation of questions --- composed by \textit{subject} and \textit{body} fields --- and answers --- a \textit{comment} field --- we adapted our system  for the 
different English subtasks:

\begin{itemize}
  \item \textbf{Subtask A (question-comment similarity ranking):} we used the aforementioned similarity-based features at three levels: question subject vs. comment, question body 
vs. comment, and full question vs. comment.
	\item \textbf{Subtask B (question-related question similarity ranking):} for this subtask we measured the similarities at body, subject, and full question level.
  \item \textbf{Subtask C (question-external comment similarity ranking):} %since this subtask requires to rank the comments of related questions respect to an original question, 
  we employed all the features of Subtasks A and B, plus the similarities of the original question --- subject, body, and full levels --- with the related question comments.
\end{itemize}

In order to rank the questions and comments, we selected a variant of Support Vector Machines (SVM)~\cite{hearst1998support} optimized for ranking problems: 
SVM$_{rank}$~\cite{joachims2002optimizing}.
In our evaluation of Section~\ref{eval}, we call our system as the combination of the acronyms of our member institutions: UH-PRHLT.

Preproscessing steps included stopword removal, lemmatization, and stemming. However, for the distributed representation and knowledge graph-based features we did not employ
stemming. These decisions were motivated for performance reasons during our prototyping.

Note that each subtask allows to submit three runs per team: primary, contrastive 1 (contr. 1), and contrastive 2 (contr. 2). We used a linear kernel and optimized the cost 
factor parameter using Bayesian optimization\footnote{We used the Spearmint toolkit: \url{https://github.com/HIPS/Spearmint}}~\cite{snoek2013bayesian}. Our three runs differ only 
in the value for that parameter and correspond with the three best --- and considerably distant --- values. In addition to the ranking, the task requires also to provide a label for each instance that reflects if the question or comment is relevant to the compared question. For each subtask we optimized a threshold to determine the relevance of each instance that is based on our predicted ranking relevance.

\section{Evaluation}
\label{eval}

%In this section we evaluated our proposed approach for the SemEval 2016 Task 3 on CQA. Given one question, the task is to rank a set of questions or comments according to 
%their similarity respect to the original question. More information about the subtaks can be found in Section~\ref{intro} or in the task overview~\cite{nakov-EtAl:2016:SemEval}.
This section presents the evaluation of the SemEval 2016 Task 3 on CQA.  Details about this task, the datasets, and the three subtasks can be found in the task 
overview~\cite{nakov-EtAl:2016:SemEval}. Note that for our system we did not use data from SemEval 2015 CQA as we did not observe gains in performance.

We compared the results of our approach with those provided by the random baseline and the Google search engine when ranking the questions and comments.\footnote{Some 
considerations about the evaluation: these subtasks employed binary classification. At testing time, \textit{Bad} and 
\textit{PotentiallyUseful} are both considered \textit{false}. The same occurs with \textit{PerfectMatch} and \textit{Relevant}, which are both considered \textit{true}. In 
addition, following the rules of the task, the employed measures used only the top 10 ranked instances.} The official 
measure of the task is the Mean Average Precision (MAP), but we included also two alternative ranking measures: Average Recall (AvgRec) and Mean Reciprocal Rank (MRR). In 
addition, we 
included four classification measures: Accuracy (acc.), Precision (P), Recall (R), and F1-measure (F1). 
%Further details about the corpus and the measures are available in the task overview~\cite{nakov-EtAl:2016:SemEval}.

\subsection{Results and Discussion}
\label{results}

\begin{table*}[!t]
  \begin{center}
      \scalebox{0.99}{
      \begin{tabular}{llccccccc}
      \toprule \multicolumn{2}{l}{\textbf{}} & \multicolumn{3}{c}{\textbf{Ranking measures}} & \multicolumn{4}{c}{\textbf{Classification measures}}\\
	 &\textbf{Model}&  \textbf{MAP} & \textbf{AvgRec} & \textbf{MRR} & \textbf{Acc.} &  \textbf{P} & \textbf{R} & \textbf{F1} \\ 
	%  & & (\textit{official measure}) & & & & & &\\
\toprule
	   \multicolumn{2}{l}{\textit{Development set results}} & & & & & & &\\
	   (a) & Random baseline & 0.456 & 0.654  & 0.535  & 0.433 & 0.344 & \textbf{0.764} & 0.475 \\
	       & Search engine   & 0.538 & 0.728  & 0.631 & n/a & n/a & n/a & n/a \\
	   \midrule 
	   (b) & UH-PRHLT (primary) & \textbf{0.632} & \textbf{0.812} & \textbf{0.725} & \textbf{0.682} & \textbf{0.526} & 0.500 & 0.513\\
	       & UH-PRHLT (contr. 1) & 0.630 & 0.811 & 0.722 & 0.672 & 0.510 & 0.545 & \textbf{0.527}\\
	       & UH-PRHLT (contr. 2) & 0.630 & 0.810 & 0.722 & 0.674 & 0.514 & 0.522 & 0.518\\
	   \bottomrule
	   \multicolumn{2}{l}{\textit{Test set results}} & & & & & & &\\
	   (a) & Random baseline & 0.528 & 0.665 & 0.587 & 0.525 & 0.452 & 0.405 & 0.428 \\
	       & Search engine   & 0.595 & 0.726 & 0.678 & n/a & n/a & n/a & n/a \\
	   \midrule 
	   (b) & UH-PRHLT (primary) & 0.674 & 0.794 & 0.770 & \textbf{0.632} & \textbf{0.556} & 0.468 & 0.508 \\
	       & UH-PRHLT (contr. 1) & \textbf{0.676} & \textbf{0.795} & \textbf{0.771} & 0.624 & 0.541 & \textbf{0.501} & \textbf{0.520} \\
	       & UH-PRHLT (contr. 2) & 0.673 & 0.793 & 0.767 & 0.630 & 0.550 & 0.491 & \textbf{0.520} \\
	   \bottomrule
      \end{tabular}
      }
      
  \end{center}
  \caption{ \label{ResTaskA} Results of \textbf{Subtask A: English Question-Comment Similarity}. (a) Baselines;  (b) proposed approach.} 
\end{table*}  

\begin{table*}[!t]
  \begin{center}
      \scalebox{0.99}{
      \begin{tabular}{llccccccc}
      \toprule \multicolumn{2}{l}{\textbf{}} & \multicolumn{3}{c}{\textbf{Ranking measures}} & \multicolumn{4}{c}{\textbf{Classification measures}}\\
	 &\textbf{Model}&  \textbf{MAP} & \textbf{AvgRec} & \textbf{MRR} & \textbf{Acc.} &  \textbf{P} & \textbf{R} & \textbf{F1} \\ 
\toprule
	   \multicolumn{2}{l}{\textit{Development set results}} & & & & & & &\\
	   (a) & Random baseline  & 0.559 & 0.732 & 0.622 & 0.488 & 0.443 & \textbf{0.766} & 0.562 \\
	       & Search engine    & 0.713 & 0.861 & 0.766 & n/a & n/a & n/a & n/a \\
	   \midrule 
	   (b) & UH-PRHLT (primary)  & \textbf{0.759} & \textbf{0.911} & \textbf{0.830} & \textbf{0.762} & \textbf{0.721} & 0.724 & \textbf{0.723} \\
	       & UH-PRHLT (contr. 1)  & 0.757 & \textbf{0.911} & \textbf{0.830} & 0.758 & 0.712 & 0.729 & 0.721 \\
	       & UH-PRHLT (contr. 2)  & 0.755 & 0.910 & 0.817 & 0.758 & 0.714 & 0.724 & 0.719 \\
	   \bottomrule
	   \multicolumn{2}{l}{\textit{Test set results}} & & & & & & &\\
	   (a) & Random baseline & 0.470 & 0.679 & 0.510 & 0.452 & 0.404 & 0.326 & 0.361 \\
	       & Search engine  & 0.747 & 0.883 & 0.838  & n/a & n/a & n/a & n/a \\
	   \midrule 
	   (b) & UH-PRHLT (primary) & 0.767 & 0.903 & 0.830 & 0.766 & 0.635 & 0.695 & 0.664\\
	       & UH-PRHLT (contr. 1) & 0.766 & 0.902 & 0.830 & 0.763 & 0.627 & \textbf{0.708} & 0.665\\
	       & UH-PRHLT (contr. 2)& \textbf{0.773} & \textbf{0.908} & \textbf{0.840} & \textbf{0.767} & \textbf{0.636} & 0.704 & \textbf{0.668}\\
	   \bottomrule
      \end{tabular}
      }
  \end{center}
  \caption{ \label{ResTaskB} Results of \textbf{Subtask B: English Question-Question Similarity}. (a) Baselines;  (b) proposed approach.} 
\end{table*}  

\begin{table*}[!t]
  \begin{center}
        \scalebox{0.99}{
      \begin{tabular}{llccccccc}
      \toprule \multicolumn{2}{l}{\textbf{}} & \multicolumn{3}{c}{\textbf{Ranking measures}} & \multicolumn{4}{c}{\textbf{Classification measures}}\\
	 &\textbf{Model}&  \textbf{MAP} & \textbf{AvgRec} & \textbf{MRR} & \textbf{Acc.} &  \textbf{P} & \textbf{R} & \textbf{F1} \\ 
\toprule
	   \multicolumn{2}{l}{\textit{Development set results}} & & & & & & &\\
	   (a) & Random baseline  & 0.138 & 0.096 & 0.160 & 0.284 & 0.070 & \textbf{0.759} & 0.128\\
	       & Search engine    & 0.306 & 0.346 & 0.360 & n/a & n/a & n/a & n/a \\
	   \midrule 
	   (b) & UH-PRHLT (primary)  & \textbf{0.383} & 0.413 & 0.420 & 0.894 & 0.242 & 0.252 & 0.247 \\
	       & UH-PRHLT (contr. 1)  & \textbf{0.383} & \textbf{0.421} & 0.425 & 0.897 & \textbf{0.252} & 0.249 & \textbf{0.250} \\
	       & UH-PRHLT (contr. 2)  & \textbf{0.383} & 0.419 & \textbf{0.435} & \textbf{0.899} & 0.251 & 0.232 & 0.241 \\
	   \bottomrule
	   \multicolumn{2}{l}{\textit{Test set results}} & & & & & & &\\
	   (a) & Random baseline & 0.150 & 0.114 & 0.152 & 0.167 & 0.296 & 0.094 & 0.143 \\
	       & Search engine   & 0.404 & 0.460 & 0.459 & n/a & n/a & n/a & n/a \\
	   \midrule 
	   (b) & UH-PRHLT (primary) & 0.432 & \textbf{0.480} & 0.478 & 0.886 & 0.376 & \textbf{0.342} & \textbf{0.359}\\
	       & UH-PRHLT (contr. 1) & \textbf{0.434} & \textbf{0.480} & \textbf{0.484} & \textbf{0.888} & \textbf{0.386} & 0.327 & 0.354\\
	       & UH-PRHLT (contr. 2) & 0.433 & \textbf{0.480} & \textbf{0.484} & \textbf{0.888} & 0.382 & 0.327 & 0.353\\
	   \bottomrule
      \end{tabular}
      }
  \end{center}
  \caption{ \label{ResTaskC} Results of \textbf{Subtask C: English Question-External Comment Similarity}. (a) Baselines;  (b) proposed approach.} 
\end{table*} 

%Too many footnotes!
The best results per partition and subtask are highlighted in bold. In 
addition, we always refer to the run with the highest performance. Finally, our percentage comparisons use always absolute values. We can see the results of Subtask A
(question-comment similarity ranking) in Table~\ref{ResTaskA}. %The performance in the development and test sets is interesting \footnote{The differences in performance are 
%similar --- for the three subtasks --- when the test set is employed. Therefore, we always refer to the development partition.} 
In terms of ranking measures, our system outperformed both the random and the search 
engine baseline.
Using the development set, we observed a MAP improvement of 9.4\% compared with the results obtained by the search engine.
We can see similar differences with respect to the other two ranking measures. 
Classification results are also superior. We obtain improvements in accuracy and F1 of 24.9\% and 5.2\% respectively. 
These results manifest the potential of the selected lexical and semantic-based features for this subtask.

Similar to Subtask A, the performance of our approach has been also superior in Subtask B (question-related question similarity ranking). As we can see in 
Table~\ref{ResTaskB}, using the development set, the improvement of MAP, AvgRec, and MRR has been of 4.6\%, 5\%, and 6.4\% respectively compared to the search engine baseline. In 
this case, the similarity between questions was easier to estimate --- also for the baselines --- and the improvements in performance were slightly reduced. With respect 
to the classification measures, we outperformed the random baseline with 27.4\% and 16.1 \% of accuracy and F1-measure respectively. 

In Table~\ref{ResTaskC} we can see the results of the Subtask C (question-external comment similarity ranking). In this case, we are ranking 100 comments (10 times more compared 
to the other subtasks). Therefore, this has been the most difficult subtask. However, we obtained improvements in line with those reported for the other 
subtasks. Compared to the search engine baseline, the MAP, AvgRec, and MRR improved 8.7\%, 8.5\%, and 7.5\% respectively when using the development partition. The accuracy 
and F1-measure improved 61.5\% and 12.2\% respectively. The largest number of comments to rank, and the use of top 10 results when measuring results, benefited our approach 
with this especially high difference in accuracy.

After the analysis of results in the three English subtasks, we highlight that the combination of 
lexical and semantic-based features that we employ in this work offers a competitive performance for the CQA task. This is true also when comparing results with other task participants. 
Our approach obtained the highest results --- with considerable margin (1.04\%) --- for subtask B. It is worth mentioning that we designed our system for the subtask B and adapted 
it later for the other tasks. However, for the other two subtasks, we obtained a low ranking position. At this point we have not discovered any coding error that could explain 
this difference. In addition, we analysed the information gain ratio of the features for the three subtasks. That results showed an average decrease of $\sim$66\% for subtasks A 
and C.
Therefore, we conclude that our approach is more adequate for tasks of similarity rather than question answering.
That analysis also manifests that the most relevant features are the word $n$-gram ones followed by the CWASA, distributed representation-based, and knowledge-graph-based ones. 
The comparison of results of all the submitted systems and task participants can be found in the task overview~\cite{nakov-EtAl:2016:SemEval}.

\section{Conclusions}
\label{conclu}

In this work we evaluated the three English subtasks of the SemEval 2016 Task 3 on CQA. In order to measure similarities, our proposed approach combined lexical and 
semantic-based features. We included simple --- and effective --- representations based on BOW, character and word $n$-grams. We also employed semantic features which 
used distributed representations of words to represent documents or to directly measure similarity by means of distributed word bidirectional alignments. The use of knowledge 
graphs generated with the BabelNet multilingual semantic network has been exploited too. Experimental results showed that our system was able to outperform --- with considerably 
differences --- the random and Google search engine baselines in all the evaluated subtasks. In addition, our approach obtained the highest results in
subtask B compared to the other task participants. This fact manifests the potential of our combination of lexical and semantic features for the CQA subtask.

As future work we will continue studying how to approach CQA with knowledge graphs and distributed representations. In addition, we will further explore how to employ this type of 
lexical and semantic-based representations for other NLP tasks such as plagiarism detection. 

 \section*{Acknowledgments}

The work of the authors of the PRHLT research center was supported by the SomEMBED TIN2015-71147-C2-1-P MINECO research project and by the Generalitat Valenciana under the grant 
ALMAMATER (PrometeoII/2014/030). 
We thank Joan Puigcerver (PRHLT) for his support and comments.

%\newpage
\bibliographystyle{naaclhlt2016} 
%\bibliography{references.bib}
\bibliography{references}

\begin{thebibliography}{}

\bibitem[\protect\citename{Baker \bgroup et al.\egroup
  }1998]{baker1998berkeley}
Collin~F Baker, Charles~J Fillmore, and John~B Lowe.
\newblock 1998.
\newblock The berkeley framenet project.
\newblock In {\em Proceedings of the 17th international conference on
  Computational linguistics-Volume 1}, pages 86--90. Association for
  Computational Linguistics.

\bibitem[\protect\citename{Blei \bgroup et al.\egroup }2003]{blei2003latent}
David~M Blei, Andrew~Y Ng, and Michael~I Jordan.
\newblock 2003.
\newblock Latent dirichlet allocation.
\newblock {\em the Journal of machine Learning research}, 3:993--1022.

\bibitem[\protect\citename{Buscaldi \bgroup et al.\egroup
  }2010]{buscaldi2010answering}
Davide Buscaldi, Paolo Rosso, Jos{\'e}~Manuel G{\'o}mez-Soriano, and Emilio
  Sanchis.
\newblock 2010.
\newblock Answering questions with an n-gram based passage retrieval engine.
\newblock {\em Journal of Intelligent Information Systems}, 34(2):113--134.

\bibitem[\protect\citename{Celikyilmaz \bgroup et al.\egroup
  }2010]{celikyilmaz2010lda}
Asli Celikyilmaz, Dilek Hakkani-Tur, and Gokhan Tur.
\newblock 2010.
\newblock Lda based similarity modeling for question answering.
\newblock In {\em Proceedings of the NAACL HLT 2010 Workshop on Semantic
  Search}, pages 1--9. Association for Computational Linguistics.

\bibitem[\protect\citename{Franco-Salvador \bgroup et al.\egroup
  }2014]{francosalvador-rosso-navigli:2014:EACL}
Marc Franco-Salvador, Paolo Rosso, and Roberto Navigli.
\newblock 2014.
\newblock A knowledge-based representation for cross-language document
  retrieval and categorization.
\newblock In {\em Proceedings of the 14th Conference of the European Chapter of
  the Association for Computational Linguistics}, pages 414--423. Association
  for Computational Linguistics.

\bibitem[\protect\citename{Franco-Salvador \bgroup et al.\egroup
  }2015a]{FrancoSalvadorCLEF2015}
Marc Franco-Salvador, Francisco Rangel, Paolo Rosso, Mariona Taul{\'e}, and
  M.~Ant{\`o}nia Mart{\'i}.
\newblock 2015a.
\newblock Language variety identification using distributed representations of
  words and documents.
\newblock In {\em Proceeding of the 6th International Conference of CLEF on
  Experimental IR meets Multilinguality, Multimodality, and Interaction (CLEF
  2015)}, volume LNCS(9283), page n/a. Springer-Verlag.

\bibitem[\protect\citename{Franco-Salvador \bgroup et al.\egroup
  }2015b]{FrancoSalvadorDSL2015}
Marc Franco-Salvador, Paolo Rosso, and Francisco Rangel.
\newblock 2015b.
\newblock Distributed representations of words and documents for discriminating
  similar languages.
\newblock In {\em Proceeding of the Joint Workshop on Language Technology for
  Closely Related Languages, Varieties and Dialects (LT4VarDial), RANLP}.

\bibitem[\protect\citename{Franco-Salvador \bgroup et al.\egroup
  }2016a]{FrancoSalvador2016:KNOSYS}
Marc Franco-Salvador, Parth Gupta, Paolo Rosso, and Rafael~E. Banchs.
\newblock 2016a.
\newblock Cross-language plagiarism detection over continuous-space
  representations of language.
\newblock {\em Pre-print submitted to journal}.

\bibitem[\protect\citename{Franco-Salvador \bgroup et al.\egroup
  }2016b]{FrancoSalvadorIPM2016}
Marc Franco-Salvador, Paolo Rosso, and Manuel~Montes y~G{\'o}mez.
\newblock 2016b.
\newblock A systematic study of knowledge graph analysis for cross-language
  plagiarism detection.
\newblock {\em Information Processing \& Management}.

\bibitem[\protect\citename{Hassan and Mihalcea}2011]{hassan2011semantic}
Samer Hassan and Rada Mihalcea.
\newblock 2011.
\newblock Semantic relatedness using salient semantic analysis.
\newblock In {\em AAAI}.

\bibitem[\protect\citename{Hearst \bgroup et al.\egroup
  }1998]{hearst1998support}
Marti~A. Hearst, Susan~T Dumais, Edgar Osman, John Platt, and Bernhard
  Scholkopf.
\newblock 1998.
\newblock Support vector machines.
\newblock {\em Intelligent Systems and their Applications, IEEE}, 13(4):18--28.

\bibitem[\protect\citename{Hou \bgroup et al.\egroup }2015]{hou2015hitszicrc}
Yongshuai Hou, Cong Tan, Xiaolong Wang, Yaoyun Zhang, Jun Xu, and Qingcai Chen.
\newblock 2015.
\newblock Hitszicrc: Exploiting classification approach for answer selection in
  community question answering.
\newblock In {\em Proceedings of the 9th International Workshop on Semantic
  Evaluation, SemEval}, volume~15 of {\em SemEval '15}, pages 196--202.
  Association for Computational Linguistics.

\bibitem[\protect\citename{Jeon \bgroup et al.\egroup }2005]{jeon2005finding}
Jiwoon Jeon, W~Bruce Croft, and Joon~Ho Lee.
\newblock 2005.
\newblock Finding similar questions in large question and answer archives.
\newblock In {\em Proceedings of the 14th ACM international conference on
  Information and knowledge management}, pages 84--90. ACM.

\bibitem[\protect\citename{Joachims}2002]{joachims2002optimizing}
Thorsten Joachims.
\newblock 2002.
\newblock Optimizing search engines using clickthrough data.
\newblock In {\em Proceedings of the eighth ACM SIGKDD international conference
  on Knowledge discovery and data mining}, pages 133--142. ACM.

\bibitem[\protect\citename{Lin and Pantel}2001]{lin2001discovery}
Dekang Lin and Patrick Pantel.
\newblock 2001.
\newblock Discovery of inference rules for question-answering.
\newblock {\em Natural Language Engineering}, 7(04):343--360.

\bibitem[\protect\citename{Mikolov \bgroup et al.\egroup
  }2013]{mikolov2013distributed}
Tomas Mikolov, Ilya Sutskever, Kai Chen, Greg~S Corrado, and Jeff Dean.
\newblock 2013.
\newblock Distributed representations of words and phrases and their
  compositionality.
\newblock In {\em Advances in Neural Information Processing Systems 26}, pages
  3111--3119.

\bibitem[\protect\citename{Nakov \bgroup et al.\egroup }2015]{nakov2015semeval}
Preslav Nakov, Llu{\'\i}s M{\`a}rquez, Walid Magdy, and Alessandro Moschitti.
\newblock 2015.
\newblock Semeval-2015 task 3: Answer selection in community question
  answering.
\newblock In {\em Proceedings of the 9th International Workshop on Semantic
  Evaluation}, SemEval '15, pages 269--281. Association for Computational
  Linguistics.

\bibitem[\protect\citename{Nakov \bgroup et al.\egroup
  }2016]{nakov-EtAl:2016:SemEval}
Preslav Nakov, Llu\'{i}s M\`{a}rquez, Walid Magdy, Alessandro Moschitti, Jim
  Glass, and Bilal Randeree.
\newblock 2016.
\newblock {SemEval}-2016 task 3: Community question answering.
\newblock In {\em Proceedings of the 10th International Workshop on Semantic
  Evaluation}, SemEval '16, San Diego, California, June. Association for
  Computational Linguistics.

\bibitem[\protect\citename{Navigli and Ponzetto}2012]{navigli2012babelnet}
Roberto Navigli and Simone~Paolo Ponzetto.
\newblock 2012.
\newblock Babelnet: The automatic construction, evaluation and application of a
  wide-coverage multilingual semantic network.
\newblock {\em Artificial Intelligence}, 193:217--250.

\bibitem[\protect\citename{Ravichandran and
  Hovy}2002]{ravichandran2002learning}
Deepak Ravichandran and Eduard Hovy.
\newblock 2002.
\newblock Learning surface text patterns for a question answering system.
\newblock In {\em Proceedings of the 40th annual meeting on association for
  computational linguistics}, pages 41--47. Association for Computational
  Linguistics.

\bibitem[\protect\citename{Rosso \bgroup et al.\egroup }2012]{rosso2012voice}
Paolo Rosso, Llu{\'\i}s-F Hurtado, Encarna Segarra, and Emilio Sanchis.
\newblock 2012.
\newblock On the voice-activated question answering.
\newblock {\em IEEE Transactions on Systems, Man, and Cybernetics--Part C},
  42(1):75--85.

\bibitem[\protect\citename{Salton and McGill}1986]{Salton:1986:IMI:576628}
Gerard Salton and Michael~J. McGill.
\newblock 1986.
\newblock {\em Introduction to Modern Information Retrieval}.
\newblock McGraw-Hill, Inc.

\bibitem[\protect\citename{Snoek}2013]{snoek2013bayesian}
Jasper Snoek.
\newblock 2013.
\newblock {\em Bayesian Optimization and Semiparametric Models with
  Applications to Assistive Technology}.
\newblock {Ph.D.} thesis, University of Toronto.

\bibitem[\protect\citename{Tran \bgroup et al.\egroup }2015]{tran2015jaist}
Quan~Hung Tran, Vu~Tran, Tu~Vu, Minh Nguyen, and Son~Bao Pham.
\newblock 2015.
\newblock Jaist: Combining multiple features for answer selection in community
  question answering.
\newblock In {\em Proceedings of the 9th International Workshop on Semantic
  Evaluation}, volume~15 of {\em SemEval '15}, pages 215--219. Association for
  Computational Linguistics.

\bibitem[\protect\citename{Voorhees}1999]{voorhees1999trec}
Ellen~M. Voorhees.
\newblock 1999.
\newblock The trec-8 question answering track report.
\newblock In {\em Trec}, volume~99, pages 77--82.

\end{thebibliography}

\end{document}